# LoRA-PT: **Lo**w-**R**ank **A**dapting UNETR for Hippocampus Segmentation Using **P**rincipal **T**ensor Singular Values and Vectors

Guanghua He[a,b], Wangang Cheng[b], Hancan Zhu[b] and Gaohang Yu[a,*]

[a] Department of Mathematics, Hangzhou Dianzi University, Hangzhou, 310018, China

[b] School of Mathematics, Physics and Information, Shaoxing University, Shaoxing, 312000, China

* Corresponding author: Gaohang Yu, E-mail address: maghyu@163.com

## ABSTRACT

The hippocampus is an important brain structure involved in various psychiatric disorders, and its automatic and accurate segmentation is vital for studying these diseases. Recently, deep learning-based methods have made significant progress in hippocampus segmentation. However, training deep neural network models requires substantial computational resources, time, and a large amount of labeled training data, which is frequently scarce in medical image segmentation. To address these issues, we propose LoRA-PT, a novel parameter-efficient fine-tuning (PEFT) method that transfers the pre-trained UNETR model from the BraTS2021 dataset to the hippocampus segmentation task. Specifically, LoRA-PT divides the parameter matrix of the transformer structure into three distinct sizes, yielding three third-order tensors. These tensors are decomposed using tensor singular value decomposition to generate low-rank tensors consisting of the principal singular values and vectors, with the remaining singular values and vectors forming the residual tensor. During fine-tuning, only the low-rank tensors (i.e., the principal tensor singular values and vectors) are updated, while the residual tensors remain unchanged. We validated the proposed method on three public hippocampus datasets, and the experimental results show that LoRA-PT outperformed state-of-the-art PEFT methods in segmentation accuracy while significantly reducing the number of parameter updates. Our source code is available at https://github.com/WangangCheng/LoRA-PT/tree/LoRA-PT.

## 1. Introduction

The hippocampus is an important brain structure involved in cognitive functions such as memory and spatial navigation. In neuroimaging research and clinical applications, morphological changes in the hippocampus are important biomarkers for neurodegenerative diseases such as Alzheimer’s disease (AD), therefore precise segmentation is required for quantitative analysis and pathological studies [1-3]. However, accurate automated segmentation of the hippocampus remains challenging owing to its complex anatomical structure, indistinct boundaries, and low contrast with surrounding tissues. These factors are further exacerbated by significant anatomical variability among individuals, particularly in pathological conditions, making it difficult for generic segmentation algorithms to perform consistently across different subjects.

Recently, deep learning methods have significantly progressed in the field of medical image segmentation [4-6]. However, these methods have limitations when applied to hippocampus segmentation. Deep learning models typically require a large volume of labeled data for training; however, annotating medical images is time-consuming and labor-intensive, making it difficult to acquire sufficient labeled data. In addition, training these models requires substantial computational resources, which poses a significant challenge for research institutions and clinical settings with limited computational capabilities [7].

Transfer learning leverages knowledge acquired from training a model in one domain to improve learning efficiency and performance in another domain while also reducing the demand for data and computational resources to some extent [8, 9]. Traditional transfer learning methods typically fine-tune all parameters of the pre-trained model (referred to "full tuning"), requiring substantial computational resources and training time and being prone to overfitting. To address these issues, parameter-efficient fine-tuning (PEFT) methods have been proposed [10]. PEFT reduces the number of parameter updates by adjusting only a small portion of the model's parameters, thereby improving fine-tuning efficiency. Recently, a PEFT method called low-rank adaptation (LoRA) has gained significant attention [11]. This method applies a low-rank approximation to the parameter matrix of the pre-trained model and updates only the low-rank portion of the parameters during fine-tuning.

Building on the success of LoRA, we aim to further optimize fine-tuning by exploiting the correlations between the parameters of different layers in deep neural network models. These models process input features layer by layer, with the parameters of each layer exhibiting strong interdependencies. For example, the UNETR model [12] contains an encoder branch with 12 sequential transformer layers, where the parameter matrices across different layers are highly correlated. When these matrices are concatenated into a tensor, the resulting tensor naturally demonstrates low-rank properties. Based on this observation, we extend LoRA to the tensor form and propose a new PEFT method based on the tensor singular value decomposition (t-SVD), called LoRA-PT. Specifically, we categorize the parameter matrices of the transformer's structure into three different sizes, forming three third-order tensors, and decompose them using the t-SVD. The decomposition splits the tensor into principal low-rank and residual parts. During the fine-tuning, we keep the residual tensor frozen and update only the low-rank tensor.

Using the proposed LoRA-PT method, we adapt the UNETR model pre-trained on the BraTS2021 dataset for hippocampus segmentation on the EADC-ADNI, LPBA40, and HFH datasets. Experimental results show that, with a small number of training samples (e.g., five samples), LoRA-PT improves the average Dice score by 0.57%, 0.72%, and 2.34%, and reduces the HD95 by 0.15 mm, 0.02 mm, and 0.75 mm on the EADC-ADNI, LPBA40, and HFH datasets, respectively, compared to full tuning. The number of updated parameters is only 3.16% of those in full tuning. In addition, LoRA-PT outperforms state-of-the-art PEFT methods, including Adapter [10], LoRA [11], SSF [13], LoTR [14], and PISSA [15], in terms of segmentation accuracy.

The main contributions of the LoRA-PT method are as follows:

- Compared to LoRA, which fine-tunes each matrix individually, LoRA-PT uses the t-SVD to decompose and fine-tune the tensor formed by the parameter matrices of different layers, effectively leveraging the overall low-rank structure of the model.
- LoRA-PT initializes the low-rank tensor using principal tensor singular values and vectors, and enhances information interaction between parameters of different layers during fine-tuning through the tensor product (t-product).

- We applied LoRA-PT to transfer the UNETR model pre-trained on the BraTS2021 dataset to hippocampus segmentation, and the results demonstrated that LoRA-PT outperformed several state-of-the-art PEFT methods.

## 2. Related work

### *2.1. Tensor decomposition*

Tensor decomposition is a fundamental tool for multi-dimensional data analysis and has been extensively applied in signal processing, machine learning, and computer vision. These techniques uncover the intrinsic structure of high-dimensional data and enhance computational efficiency by representing tensors in lower-dimensional subspaces. Classic decomposition frameworks include CANDECOMP/PARAFAC (CP), Tucker decomposition, tensor train (TT), and tensor singular value decomposition (t-SVD) [16, 17]. CP decomposition approximates a tensor as a sum of rank-one components and is particularly suitable for multi-channel signal processing and chemometric analysis [18]. Tucker decomposition represents a tensor through a core tensor and a set of factor matrices, allowing flexible modeling of spatial and temporal dependencies in tasks such as image and video analysis [19]. TT decomposition encodes high-order tensors as a chain of third-order tensors, enabling significant reductions in memory and computation [20].

Among these, t-SVD has emerged as a powerful method for third-order tensor decomposition. It provides an optimal low-rank approximation under the Frobenius norm while preserving orthogonality and enabling matrix-like algebraic operations in the tensor domain [21]. Kilmer et al. [22] further demonstrated that high-dimensional datasets can be more efficiently represented using tensor-tensor algebra, and established optimality results for truncated t-SVDs. Recent efforts have focused on extending t-SVD to higher-order tensors. Qin et al. [23] generalized the t-SVD formulation to arbitrary order-$d$ tensors via unfolding and reshaping operations, demonstrating strong performance in high-dimensional image completion. Liu et al. [24] revisited t-SVD from a basic-element perspective and introduced a higher-order convolution operator to enable consistent decomposition across arbitrary-order tensors. To further enhance scalability and robustness, Qin et al. [25] proposed randomized low-rank approximations under the t-SVD framework, enabling efficient recovery of corrupted visual data.

In parallel, several studies have improved the optimization objectives associated with t-SVD. Zhang et al. [26] introduced a novel non-convex minimax concave penalty function that better preserves significant singular components, improving tensor recovery under severe corruption. Zheng et al. [27] proposed a tensor n-tubal rank by compressing high-order tensors into third-order forms via mode-grouping, enabling effective low-rank modeling through standard t-SVD. Additional studies have explored transform-based t-SVD extensions, such as multiplex transformed tensor decomposition [28] and mode-$k$ tensor decomposition strategies [29], which generalize the t-product to operate along arbitrary modes. These approaches offer more flexible and adaptive decomposition schemes for multidimensional image recovery.

### *2.2. Deep learning for hippocampus segmentation*

Deep learning has significantly advanced hippocampus segmentation in recent years. Lin et al. proposed a three-dimensional (3D) multi-scale attention U-Net that enhanced segmentation accuracy and robustness by integrating multi-scale features and attention mechanisms [30]. Liu et al. developed a multi-model convolutional neural network (CNN) for simultaneous hippocampus segmentation and AD classification [6]. Shi et al. introduced a dual dense context-aware network, integrating dense features and contextual information to improve segmentation accuracy and robustness [31]. Yang et al. proposed a multi-scale CNN toolkit for automated hippocampal subfield segmentation (CAST), achieving high precision through hierarchical feature extraction [32]. Goubran et al. designed a 3D CNN tailored for brains with severe atrophy, addressing the issues of conventional methods under pathological conditions [33]. Zhu et al. proposed a dilated dense U-Net for infant hippocampal subfield segmentation, which improved the recognition of fine and complex anatomical structures [34]. He et al. introduced a deep CNN with a boundary refinement module, significantly enhancing the delineation of hippocampal edges [4].

In addition, transfer learning has been applied to hippocampus segmentation to improve model adaptability and generalization. Ataloglou et al. proposed a rapid and accurate hippocampus segmentation method by combining CNN ensembles with transfer learning [35]. Balboni et al. investigated the impact of transfer learning on 3D CNNs for segmenting hippocampi in patients with mild cognitive impairment and AD, demonstrating improved accuracy and efficiency through enhanced adaptability to disease-specific changes, thereby supporting early AD diagnosis [36]. Opbroek et al. introduced a feature space transformation method that mapped features from one scanner's domain to another, addressing inconsistencies in image characteristics across different imaging devices [37].

### *2.3. Parameter-efficient fine-tuning*

In contrast to traditional transfer learning, PEFT methods adapt pre-trained models by adjusting only a small subset of their parameters. This significantly reduces the reliance on large amounts of labeled data and computational resources. Houlsby et al. introduced the adapter module, embedding a small number of trainable parameters into the pre-trained transformer model, allowing it to adapt to new tasks while maintaining most parameters fixed [10]. Lian et al. proposed SSF, which incorporated learnable scaling and rotation factors after each transformer operation, updating only the newly introduced parameters and freezing the rest of the model [13]. Mou et al. proposed T2i-adapter, specifically designed for text-to-image diffusion models to enhance generative controllability [38]. Zhang et al. developed LLaMA-Adapter, leveraging a zero-initialized attention mechanism for efficient adaptation [39]. Li and Liang proposed prefix-tuning, which optimized a small set of prefix tokens prepended to the input, significantly reducing the fine-tuning cost [40].

Hu et al. developed LoRA, which fine-tuned pre-trained models by applying low-rank matrix decomposition to their weight matrices while preserving the original model structure [11]. Chen et al. introduced SuperLoRA, a new PEFT framework that optimized multi-layer attention modules [41]. It unified and extended most existing LoRA variants to enhance model performance. Liu et al. introduced DoRA, a PEFT method based on weight decomposition and low-rank adaptation [42]. Si et al. presented FLoRA, which used a low-rank core space for PEFT, preserving the original dimensional structure [43]. Kopiczko et al. proposed Vera, which enhanced model performance by learning small scaling vectors while freezing random matrices in the model architecture [44]. Bershatsky et al. introduced LoTR, which performed tensor low-rank adaptation using TT decomposition to optimize model weights [14]. Jie and Deng proposed Fact, a lightweight adaptation method for vision transformer models that selectively adjusted key factors using TT and Tucker decomposition [45]. Meng et al. developed PISSA, which enabled efficient fine-tuning by adapting the principal singular values and vectors of large language models [15].

The previously mentioned PEFT methods are listed in Table 1. As can be observed, 1) most current PEFT methods are based on matrix decomposition, although some tensor decomposition methods, including TT and Tucker decompositions, have been proposed in recent years. We are the first to introduce a t-SVD-based PEFT method, which effectively uses the t-product for inter-layer information exchange. 2) The majority of existing PEFT methods focus on natural language processing (NLP) and CV tasks, with very few methods specifically designed for medical image analysis (MIA). This highlights the novelty of our proposed t-SVD-based PEFT method, which provides a promising solution for fine-tuning models in medical imaging applications.

Table 1. Comparison of representative parameter-efficient fine-tuning (PEFT) methods by decomposition type, application domain, and key characteristics.

| Method | Year | Decomposition Type | Application Domain | Key Characteristics |
|---|---|---|---|---|
| Adapter [10] | 2019 | / | NLP | Inserts lightweight modules for task-specific adaptation with minimal parameter updates. |
| Prefix-Tuning [40] | 2021 | / | NLP | Tunes prefix token parameters to adapt pre-trained models without altering core weights. |
| SSF [13] | 2022 | / | CV | Adds scaling and rotation parameters, training only these. |
| LoRA [11] | 2022 | Matrix | NLP | Applies low-rank decomposition to reduce the number of trainable parameters. |
| PISSA [15] | 2024 | Matrix | NLP | Adapts principal singular components for efficient fine-tuning. |
| SuperLoRA [41] | 2024 | Matrix/Tensor | CV | Unifies and extends existing LoRA variants within a generalized PEFT framework. |
| FLoRA [43] | 2024 | Tensor (Tucker) | NLP/CV | Applies Tucker decomposition to preserve dimensional structure in low-rank PEFT. |
| DoRA [42] | 2024 | Matrix | NLP/CV | Decomposes weights into magnitude and direction for low-rank adaptation. |
| Vera [44] | 2023 | Matrix | NLP/CV | Employs vector-based random matrix adaptation to improve model performance. |
| Fact [45] | 2023 | Tensor (TT/Tucker) | CV | Selectively tunes tensor factors using TT/Tucker decomposition. |
| LoTR [14] | 2024 | Tensor (TT) | CV | Uses TT decomposition for low-rank tensor adaptation. |
| **LoRA-PT (Ours)** | **2025** | **Tensor (t-SVD)** | **MIA** | **t-SVD based PEFT for medical image segmentation** |

## 3. Methods

UNETR consists of an encoder and a decoder. The encoder is constructed by stacking 12 transformers, while the decoder comprises convolutional layers, batch normalization, and ReLU activation functions. These components are connected via a long skip connection that includes deconvolution, convolution, batch normalization, and ReLU activation functions. For a detailed description of the UNETR architecture, refer to the original paper [12]. Because most of the parameters in UNETR are concentrated in the transformer structure, the proposed LoRA-PT method fine-tuned the transformer parameters, keeps the decoder parameters fully updated, and froze the parameters in the long skip connection. The framework of LoRA-PT, shown in Fig. 1, consists of three components: tensorization of UNETR, t-SVD of tensors, and extraction and fine-tuning of principal tensor singular values and vectors.

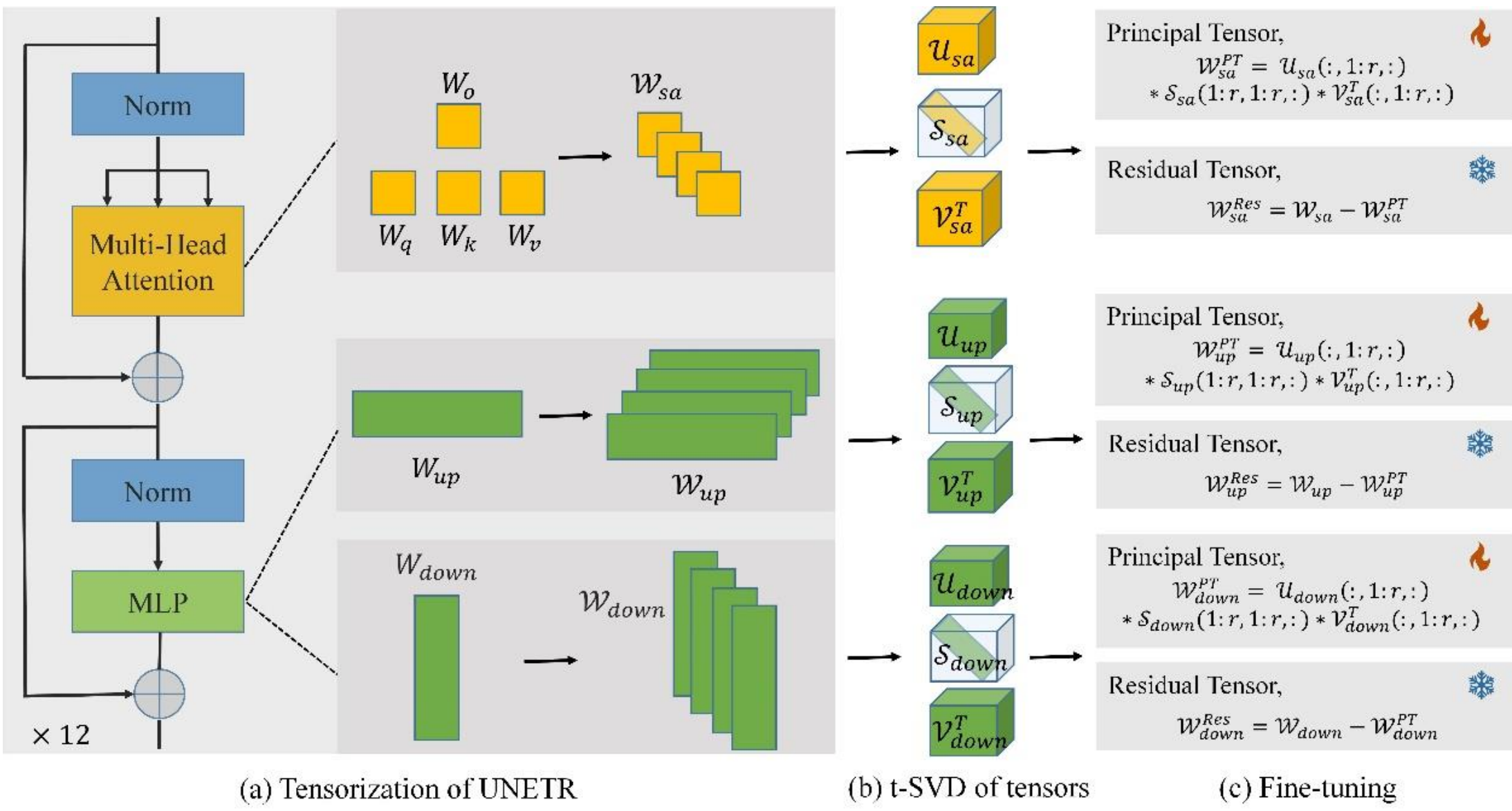

Fig. 1. Schematic of the LoRA-PT method framework for parameter-efficient fine-tuning. (a) The encoder of UNETR is tensorized into three distinct tensors. (b) These tensors are decomposed using t-SVD. (c) Principal tensor singular values and vectors are extracted and fine-tuned.

### *3.1. Tensorization of UNETR*

The transformer layers in UNETR consisted of two primary modules: multi-head self-attention (MHSA) and multi-layer perceptron (MLP). In the MHSA module, the input $X \in R^{n\times d}$ is linearly projected to query, key, and value representations using learnable weight matrices $W_q, W_k, and\ W_v \in R^{d\times d}$, respectively, and the output is transformed using $W_o \in R^{d\times d}$. The MHSA mechanism divides these transformations into $N_h$ attention heads. For the $i$-th head, the corresponding projection matrices are $W_q^{(i)}$, $W_k^{(i)}$, $W_v^{(i)}$, $W_o^{(i)} \in R^{d\times d_h}$, where $d_h = d/N_h$ is the dimensionality of each head. The MHSA output is computed as follows:

$$MHSA(X) = \sum_{i=1}^{N_h} softmax(\frac{XW_q^{(i)}(XW_k^{(i)})^T}{\sqrt{d_h}})XW_v^{(i)}(W_o^{(i)})^T.$$

The MLP module consists of two fully connected (FC) layers. Ignoring bias terms for simplicity, the MLP operation is defined as follows:

$$MLP(X) = GELU(XW_{up})W_{down},$$

where $W_{up} \in R^{d\times 4d}$ and $W_{down} \in R^{4d\times d}$ represent the weights of the two FC layers. The GELU activation function is applied between the layers to introduce non-linearity [46].

Based on the above analysis, each transformer layer contained four $d \times d$ matrices in the MHSA module and two weight matrices in the MLP module, one of size $d \times 4d$ and other of $4d \times d$. Because UNETR included 12 transformer layers, the complete model comprised 48 $d \times d$ matrices, 12 $d \times 4d$ matrices, and 12 $4d \times d$ matrices. As shown in Fig. 1(a), these matrices were stacked along the third mode to construct three third-order tensors. Specifically, the 48 $d \times d$ matrices form $\mathcal{W}_{sa} \in R^{d\times d\times 48}$, 12 $d \times 4d$ matrices form $\mathcal{W}_{up} \in R^{d\times 4d\times 12}$, and 12 $4d \times d$ matrices form $\mathcal{W}_{down} \in R^{4d\times d\times 12}$.

### *3.2. T-SVD of Tensors*

After tensorizing the UNETR parameters, the next step is to decompose these tensors to enable efficient fine-tuning. The t-SVD is a powerful generalization of matrix SVD to third-order tensors, decomposing a tensor into three simpler components and effectively capturing multi-dimensional correlations in a compact, structured form. To help readers unfamiliar with tensor algebra, we first introduce the essential tensor concepts, followed by the definition of block circulant matrices and the t-product. Finally, we present the formal definition of t-SVD [21].

#### *3.2.1. Basic tensor concepts*

Consider $\mathcal{A} \in R^{n_1\times n_2\times n_3}$, a third-order tensor represented by a 3D data cube with three modes. By fixing one of the mode indices, the tensor can be sliced along that mode to produce different types of two-dimensional (2D) matrices, commonly referred to as tensor slices. Among the various types of tensor slices, the most frequently used in t-SVD are the frontal slices, which are obtained by fixing the index of the third mode. Each frontal slice is a 2D matrix denoted by $A_i = \mathcal{A}(:,:,i)$, where $i = 1,2,\dots,n_3$. In addition to frontal slices, two other types of slices can be defined by assigning different mode indices. To obtain lateral slices, fix the index of the second mode: $\mathcal{A}(:,i,:)$, where $i = 1,2,\dots,n_2$, and to obtain horizontal slices, fix the index of the first mode: $\mathcal{A}(i,:,:)$, where $i = 1,2,\dots,n_1$.

We introduced several important tensor-related definitions used in t-SVD. A third-order tensor $\mathcal{A} \in R^{n_1\times n_2\times n_3}$ is said to be *f-diagonal* if each of its frontal slices $A_i$ is a diagonal matrix. The *transpose* of the tensor $\mathcal{A}$, denoted $\mathcal{A}^T \in R^{n_2\times n_1\times n_3}$, is obtained by transposing each frontal slice and reversing the order of slices from the second to the last. The *MatVec* operation rearranges the frontal slices of a tensor into a tall matrix by stacking them vertically as follows:

$$\mathrm{MatVec}(\mathcal{A}) = \begin{bmatrix} A_1 \\ A_2 \\ \vdots \\ A_{n_3} \end{bmatrix}.$$

The *fold* operation is the inverse of MatVec, reconstructing the original tensor from its stacked matrix form. That is,

$$\mathrm{fold}\big(\mathrm{MatVec}(\mathcal{A})\big) = \mathcal{A}.$$

#### *3.2.2. Block circulant matrices*

A circulant matrix is a square matrix in which each column is a downward cyclic shift of the previous column. For example, consider a vector $a = [a_1, a_2, a_3]^T$, its corresponding circulant matrix is expressed as follows:

$$\text{circ}(a) = \begin{bmatrix} a_1 & a_3 & a_2 \\ a_2 & a_1 & a_3 \\ a_3 & a_2 & a_1 \end{bmatrix}.$$

An important property of circulant matrices is that they are diagonalizable by the discrete Fourier transform (DFT). Specifically, for any vector $v \in R^n$, let $F_n$ be the $n \times n$ DFT matrix and $F_n^*$ its conjugate transpose. Subsequently, the matrix $F_n \text{circ}(v) F_n^*$ is diagonal.

This concept naturally extends to block circulant matrices, which are widely used in tensor computations. Given a third-order tensor $\mathcal{A} \in R^{n_1 \times n_2 \times n_3}$, its block circulant matrix, denoted as $\text{circ}(\mathcal{A})$, is constructed by arranging the frontal slices $A_i = \mathcal{A}(:,:,i), i = 1,2,\dots,n_3$ in a circulant structure as follows:

$$\text{circ}(\mathcal{A}) = \begin{bmatrix} A_1 & A_{n_3} & \cdots & A_2 \\ A_2 & A_1 & \cdots & A_3 \\ \vdots & \vdots & \vdots & \vdots \\ A_{n_3} & A_{n_3-1} & \cdots & A_1 \end{bmatrix}.$$

Similarly, block circulant matrices can also be block-diagonalized using the DFT. Let $F_{n_3}$ be the $n_3 \times n_3$ DFT matrix, and $I_{n_1}$ and $I_{n_2}$ be identity matrices of size $n_1 \times n_1$ and $n_2 \times n_2$, respectively. Subsequently, the following block-diagonalization holds [21, 47]:

$$\left(F_{n_3} \otimes I_{n_1}\right) \text{circ}(\mathcal{A}) \left(F_{n_3}^* \otimes I_{n_2}\right) = \begin{bmatrix} \hat{A}_1 & 0 & \cdots & 0 \\ 0 & \hat{A}_2 & \cdots & 0 \\ \vdots & \vdots & \vdots & \vdots \\ 0 & 0 & \cdots & \hat{A}_{n_3} \end{bmatrix},$$

where $\otimes$ denotes the Kronecker product, and $\hat{A}_i$ denotes the frontal slices of the tensor $\hat{\mathcal{A}}$, which is obtained by applying the DFT along the third mode of $\mathcal{A}$. This block-diagonalization provides the foundation for the efficient computation of the t-product, as detailed in the subsequent subsection.

### *3.2.3. T-product*

The t-product defines a multiplication operation between two third-order tensors, generalizing the standard matrix multiplication to the tensor domain.

**Definition 1 (t-product):** Let $\mathcal{A} \in R^{n_1 \times n_2 \times n_3}$ and $\mathcal{B} \in R^{n_2 \times l \times n_3}$ be two third-order tensors. Their t-product $\mathcal{C} = \mathcal{A} * \mathcal{B}$ is a tensor of size of $n_1 \times l \times n_3$, defined as follows:

$$\mathcal{C} = \text{fold}\left(\text{circ}(\mathcal{A}) \text{MatVec}(\mathcal{B})\right).$$

We have the following identity:

$$\text{circ}(\mathcal{A}) \text{MatVec}(\mathcal{B}) = (F_{n_3}^* \otimes I_{n_1})((F_{n_3} \otimes I_{n_1}) \text{circ}(\mathcal{A}) (F_{n_3}^* \otimes I_{n_2}))(F_{n_3} \otimes I_{n_2}) \text{MatVec}(\mathcal{B}).$$

Based on the DFT-based block-diagonalization of block circulant matrices, this identity enables an efficient implementation of the t-product $\mathcal{C} = \mathcal{A} * \mathcal{B}$, which proceeds as follows: (1) The fast Fourier transform (FFT) is applied along the third mode of both $\mathcal{A}$ and $\mathcal{B}$ to obtain $\hat{\mathcal{A}}$ and $\hat{\mathcal{B}}$; (2) Matrix multiplication is performed on each pair of corresponding frontal slices in the Fourier domain: $\hat{\mathcal{C}}(:,:,i) = \hat{\mathcal{A}}(:,:,i) \hat{\mathcal{B}}(:,:,i),\ i = 1,2,\dots,n_3$; (3) The inverse FFT (IFFT) is applied along the third mode of $\hat{\mathcal{C}}$ to reconstruct the final result: $\mathcal{C}$=ifft($\hat{\mathcal{C}}$, axis=3).

We now introduce two important tensor structures: the *identity* and *orthogonal tensors*. The identity tensor $\mathcal{I} \in R^{n \times n \times n_3}$ is defined such that its first frontal slice is the $n \times n$ identity matrix, and all remaining frontal slices are zero matrices. It serves as the multiplicative identity under the t-product, such that for any tensor $\mathcal{A} \in R^{n \times n \times n_3}$, it satisfies

$$\mathcal{A} * \mathcal{I} = \mathcal{I} * \mathcal{A} = \mathcal{A}.$$

A tensor $\mathcal{Q} \in R^{n\times n\times n_3}$ is said to be orthogonal if it satisfies $\mathcal{Q} * \mathcal{Q}^T = \mathcal{Q}^T * \mathcal{Q} = \mathcal{I}$, where $\mathcal{Q}^T$ denotes the transpose of $\mathcal{Q}$.

*3.2.4. T-SVD*

With the t-product and related definitions, we now introduce the t-SVD [21].

**Definition 2 (t-SVD)**: Let $\mathcal{A} \in R^{n_1\times n_2\times n_3}$ be a third-order tensor. Subsequently, $\mathcal{A}$ can be decomposed as follows:

$$\mathcal{A} = \mathcal{U} * \mathcal{S} * \mathcal{V}^T,$$

where $\mathcal{U}$ and $\mathcal{V}$ represent orthogonal tensors with the size of $n_1 \times n_1 \times n_3$ and $n_2 \times n_2 \times n_3$, respectively, $\mathcal{S}$ represents an f-diagonal tensor with the size of $n_1 \times n_2 \times n_3$, and $\mathcal{V}^T$ is the transpose of $\mathcal{V}$. The t-SVD can be efficiently computed using the FFT, as outlined in Algorithm 1.

---

**Algorithm 1:** t-SVD of a third-order tensor

**Input**: A tensor $\mathcal{A} \in R^{n_1\times n_2\times n_3}$

**Output**: $\mathcal{U}$, $\mathcal{S}$, $\mathcal{V}$ such that $\mathcal{A} = \mathcal{U} * \mathcal{S} * \mathcal{V}^T$

Apply FFT along the third mode of $\mathcal{A}$: $\hat{\mathcal{A}}$=fft($\mathcal{A}$, axis=3)

For $i = 1$ to $n_3$

$[U, S, V] = \mathrm{svd}(\hat{\mathcal{A}}(:,:,i))$

$\hat{\mathcal{U}}(:,:,i) = U$; $\hat{\mathcal{S}}(:,:,i) = S$; $\hat{\mathcal{V}}(:,:,i) = V$

$\mathcal{U}$=ifft($\hat{\mathcal{U}}$, axis=3); $\mathcal{S}$=ifft($\hat{\mathcal{S}}$, axis=3); $\mathcal{V}$=ifft($\hat{\mathcal{V}}$, axis=3)

---

*3.3. Extraction and fine-tuning of principal tensor singular values and vectors*

As shown in Fig. 1(b), we performed the t-SVD on the three concatenated parameter tensors $\mathcal{W}_{sa}$, $\mathcal{W}_{up}$, and $\mathcal{W}_{down}$, as follows:

$$\mathcal{W}_{sa} = \mathcal{U}_{sa} * \mathcal{S}_{sa} * \mathcal{V}_{sa}^T,$$

$$\mathcal{W}_{up} = \mathcal{U}_{up} * \mathcal{S}_{up} * \mathcal{V}_{up}^T,$$

$$\mathcal{W}_{down} = \mathcal{U}_{down} * \mathcal{S}_{down} * \mathcal{V}_{down}^T,$$

where

$$\mathcal{U}_{sa} \in R^{d\times d\times 48}, \mathcal{S}_{sa} \in R^{d\times d\times 48}, \mathcal{V}_{sa} \in R^{d\times d\times 48},$$

$$\mathcal{U}_{up} \in R^{d\times d\times 12}, \mathcal{S}_{up} \in R^{d\times 4d\times 12}, \mathcal{V}_{up} \in R^{4d\times 4d\times 12},$$

$$\mathcal{U}_{down} \in R^{4d\times 4d\times 12}, \mathcal{S}_{down} \in R^{4d\times d\times 12}, \mathcal{V}_{down} \in R^{d\times d\times 12}.$$

Using the results of the t-SVD, we extracted $r$ principal tensor singular values and vectors, as shown in Fig. 1(c).

Considering $\mathcal{W}_{sa}$ as an example, the left $r$ principal tensor singular vectors are $\mathcal{U}_{sa}(:,1:r,:)$, the right $r$ principal tensor singular vectors are $\mathcal{V}_{sa}(:,1:r,:)$, and the $r$ principal tensor singular values are $\mathcal{S}_{sa}(1:r,1:r,:)$. We then represent tensor $\mathcal{W}_{sa}$ as the sum of a principal low-rank tensor and a residual tensor as follows:

$$\mathcal{W}_{sa} = \mathcal{W}_{sa}^{PT} + \mathcal{W}_{sa}^{Res},$$

where $\mathcal{W}_{sa}^{PT}$ denotes the principal low-rank tensor, obtained by the t-product of the principal tensor singular values and vectors as follows:

$$\mathcal{W}_{sa}^{PT} = \mathcal{U}_{sa}(:,1:r,:) * \mathcal{S}_{sa}(1:r,1:r,:) * \mathcal{V}_{sa}^T(:,1:r,:),$$

and, $\mathcal{W}_{sa}^{Res} = \mathcal{W}_{sa} - \mathcal{W}_{sa}^{PT}$ denotes the residual tensor.

Under the t-SVD framework, it has been theoretically established [22] that the truncated approximation $\mathcal{W}_{sa}^{PT}$ provides an optimal low-rank approximation of $\mathcal{W}_{sa}$ under the Frobenius norm. Specifically, the approximation error is given by:

$$\|\mathcal{W}_{sa} - \mathcal{W}_{sa}^{PT}\|_F = \left(\sum_{i=r+1}^{l} \|\mathcal{S}_{sa}(i,i,:)\|_2^2\right)^{1/2},$$

where, $l$ is the tensor tubal rank of $\mathcal{W}_{sa}$. This formulation shows that the approximation error is solely determined by the discarded singular tubes beyond the rank threshold $r$, highlighting the efficiency of t-SVD in preserving the most significant structural information within the top-$r$ components.

As shown in Fig. 2, during fine-tuning, the residual tensor $\mathcal{W}_{sa}^{Res}$ remains frozen, while the principal tensor singular values and vectors $\mathcal{U}_{sa}(:,1{:}r,:)$, $\mathcal{S}_{sa}(1{:}r,1{:}r,:)$, and $\mathcal{V}_{sa}(:,1{:}r,:)$ are updated. To enhance memory efficiency during fine-tuning, we avoid storing the full low-rank tensor $\mathcal{W}_{sa}^{PT}$. Instead, only the principal components are retained. Before inference, $\mathcal{W}_{sa}^{PT}$ is reconstructed and merged with the frozen $\mathcal{W}_{sa}^{Res}$, so that the inference process remains computationally efficient.

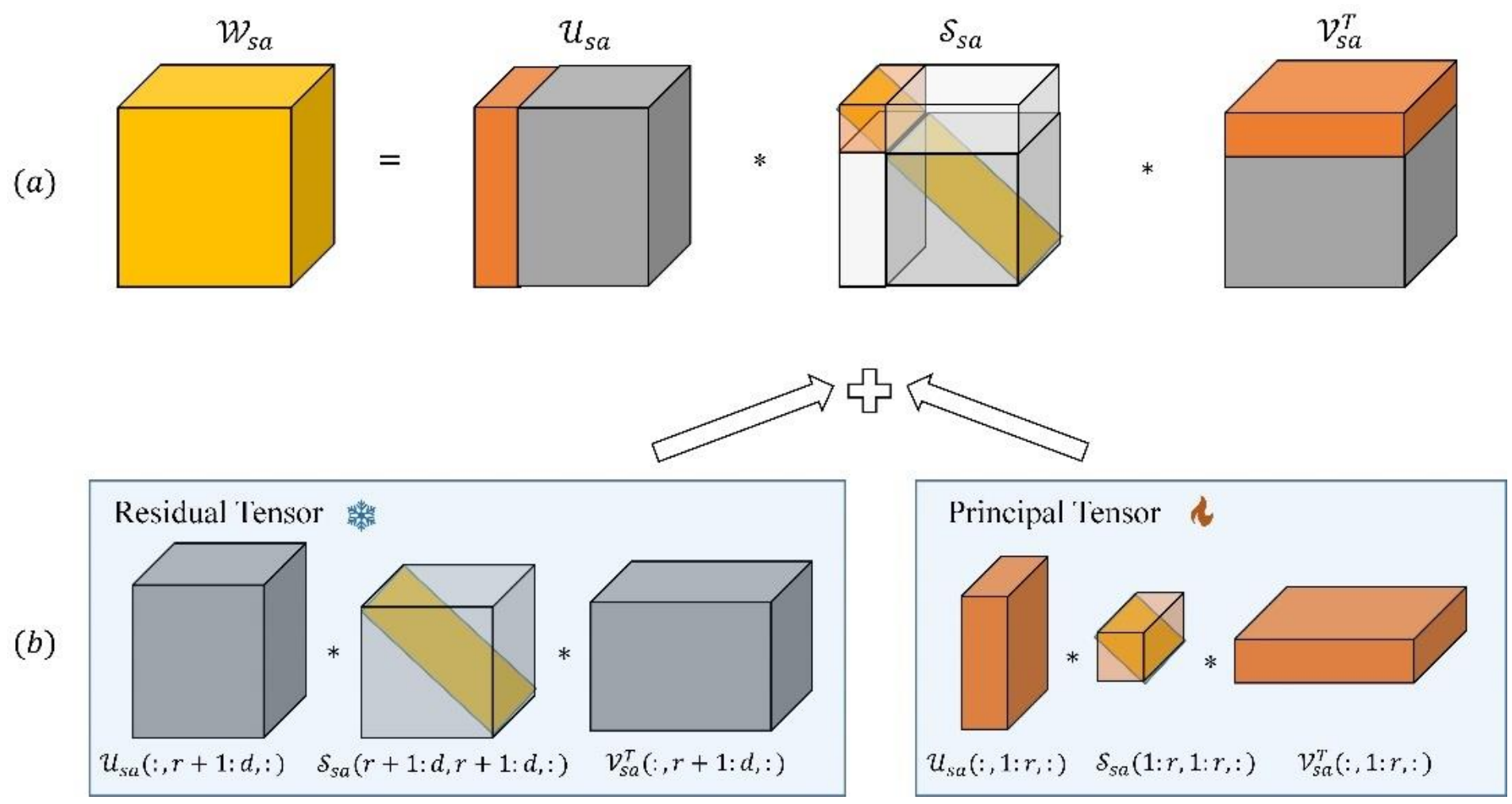

Fig. 2. (a) Tensor singular value decomposition (T-SVD) and (b) Its application in LoRA-PT: residual components frozen and principal components updated during fine-tuning.

## 4. Experiments

### 4.1. Datasets

#### 4.1.1. The Dataset for pre-training

The UNETR model was first pre-trained on the BraTS2021 dataset [48], which consisted of multi-modal magnetic resonance imaging (MRI) data, including four imaging modalities: T1, T1ce, T2, and FLAIR. The imaging data for each patient had a resolution of 240 × 240 × 155, with a voxel size of 1 × 1 × 1 mm$^3$. Most MRI scans in the BraTS2021 dataset were acquired using 3T scanners, yielding high-field MR images that are well-suited for detailed brain tumor analysis. The dataset included comprehensive tumor region annotations for 1,251 cases, with annotation categories such as enhancing tumor (ET), peritumoral edema/infiltrating tissue (ED), and necrotic and non-enhancing tumor core (NCR/NET).

### *4.1.2. Hippocampus datasets*

We applied LoRA-PT to transfer the pre-trained UNETR model to hippocampus segmentation and validated it on three hippocampus datasets: EADC-ADNI [49], LPBA40 [50], and HFH [51].

*The EADC-ADNI dataset*, obtained from the ADNI database (http://adni.loni.usc.edu/), comprises T1-weighted MR images from 135 patients, with a resolution of 197 × 233 × 189 and a voxel size of 1 × 1 × 1 $mm^3$. These images were acquired using both 1.5T and 3T MRI scanners. Hippocampus annotations were generated following the harmonized segmentation protocol developed by the European Alzheimer's Disease Consortium and Alzheimer's Disease Neuroimaging Initiative (EADC-ADNI) (www.hippocampal-protocol.net) [49]. After manual inspection, five cases were excluded owing to mismatches between the hippocampus annotations and the corresponding MR images.

*The LPBA40 dataset*, developed by the Laboratory of Neuro Imaging (LONI) in Los Angeles, is a widely used benchmark in neuroimaging research. It consists of 40 T1-weighted 3D spoiled gradient recalled echo (SPGR) scans acquired in the coronal plane using a 1.5T general electric scanner. Each scan has a slice thickness of 1.5 mm, a resolution of 256 × 124 × 256, and a voxel size of 0.8938 × 1.500 × 0.8594 $mm^3$. All images are meticulously annotated, covering 56 distinct brain tissues and anatomical structures. In this study, we focused on the hippocampal region to assess the performance of our proposed method.

*The HFH dataset* includes 50 T1-weighted MR brain images from two different MRI vendors, with varying resolutions and contrasts. The image dimensions are 256 × 124 × 256 and 512 × 124 × 512, with voxel sizes of 0.781 × 2.000 × 0.781 $mm^3$ and 0.39 × 2.00 × 0.39 $mm^3$. These images were acquired using two different MRI systems with varying field strengths. Notably, 20 images were acquired using a newer 3T MRI scanner, providing higher resolution and enhanced contrast. Because hippocampus annotations were available for only 25 images, our experiments were conducted using this annotated subset.

For all three hippocampus datasets, we applied a standardized preprocessing pipeline. Skull stripping was conducted using SynthStrip, a fully automated deep learning-based tool integrated into FreeSurfer (version 7.4.1), which required no manual intervention and ensured consistency and efficiency across subjects [52]. After skull stripping, all images were registered to the MNI152 standard space using the *flirt* command from the FSL toolbox with a 12-degree-of-freedom affine transformation [53]. To maintain precise alignment between segmentation labels and their corresponding images, the same transformation matrices were applied to the label maps, and nearest-neighbor interpolation was used during resampling to preserve label discreteness. Finally, all images and labels were resampled to an isotropic voxel size of 1 × 1 × 1 $mm^3$ and adjusted to the standard MNI152 resolution of 182 × 218 × 182.

### *4.2. Evaluation metrics*

We selected the Dice coefficient and Hausdorff Distance 95% (HD95) as evaluation metrics. The Dice coefficient measures the relative overlap volume between the automatic segmentation and the ground truth, defined as follows:

$$Dice = 2\frac{V(A \cap B)}{V(A) + V(B)},$$

where $A$ represents the ground truth, $B$ represents the automatic segmentations, and $V(S)$ represents the volume of $S$. HD95 is a robust version of the Hausdorff distance, used to evaluate the robustness of the segmentation structure and the consistency of the segmentation boundaries, defined as follows:

$$HD95(A,B) = max(h_{95}(A,B), h_{95}(B,A)),$$

where $h_{95}(A,B) = K^{95}_{a\in A} \min_{b\in B} d(a,b)$ is the 95th ranked minimum Euclidean distance between boundary points in $A$ and $B$.

*4.3. Implementation details*

We implemented the UNETR model in PyTorch and trained it on two NVIDIA GeForce RTX 4090D GPUs, preserving the original network architecture [12]. During pre-training, the T1ce modality from the BraTS2021 dataset was used as input. The Adam optimizer was used, with a cosine annealing learning rate schedule (initial learning rate of 0.001, weight decay of 1e-4, and batch size of 4). To facilitate transfer learning for hippocampus segmentation, the three annotated tumor subregions in BraTS (NCR/NET, ED, and ET) were merged into a single tumor region to formulate a binary segmentation task. A total of 1,251 annotated cases were divided into training, validation, and test sets (1000, 125, and 126 cases, respectively), following an approximate 8:1:1 ratio.

According to the standard BraTS preprocessing pipeline, we applied a range of data augmentation techniques to improve model generalization. Specifically, input images were randomly cropped into 128 × 128 × 128 cubic patches and independently flipped along each spatial axis with a probability of 50%. In addition, random intensity perturbations were introduced by applying per-channel, depth-wise scaling and shifting, simulating variations in scanning conditions and enhancing robustness to shifts in intensity distributions.

During the fine-tuning, we used the Adam optimizer with a batch size of four and adopted a polynomial learning rate schedule. The initial learning rate was set to 0.001 and decayed with a power of 0.9 at each iteration. For the network input, image patches of size 128 × 128 × 128 were extracted through random cropping. A series of data augmentation strategies were applied, including: (1) random mirroring along the axial, coronal, and sagittal planes with a probability of 0.5; (2) random intensity shifts by adding or subtracting a value within the range [−0.1, 0.1] to each voxel's intensity; and (3) random scaling of the image within a range of [0.9, 1.1]. The network was optimized using the Dice loss function, defined as follows:

$$\mathcal{L}(Y,\tilde{Y}) = -\frac{1}{N}\sum_{n=1}^{N}\frac{2Y_n\tilde{Y}_n}{Y_n+\tilde{Y}_n},$$

where $Y_n$ and $\tilde{Y}_n$ represent the ground truth and predicted probability, respectively, and $N$ represents the batch size. To mitigate overfitting, L2 regularization was applied with a weight decay of $10^{-5}$. The network was trained for a total of 1000 epochs.

During the testing phase, image patches of size 128 × 128 × 128 were extracted using a non-overlapping sliding window strategy and fed into the trained model for segmentation. To enhance prediction stability, the outputs from the last four training epochs were averaged to obtain the final inference result. In the post-processing stage, false positive hippocampal regions were removed through the following steps: (1) A binary mask was generated to identify all connected components; (2) a volumetric threshold of 1,000 $mm^3$ was applied; and (3) all connected regions smaller than this threshold were eliminated by relabeling them as background.

## 5. Results

We compared the proposed LoRA-PT method with several state-of-the-art fine-tuning methods, including full tuning, linear-probing, LoRA [11], Adapter [10], SSF [13], LoTR [14], and PISSA [15]. Linear-probing updates only the linear layers within transformer blocks. LoRA, PISSA, and LoTR represent matrix- or tensor-based adaptation methods, whereas Adapter and SSF follow alternative parameter-efficient strategies. A detailed introduction to these methods is provided in Section 2.

### *5.1. Selection of optimal rank*

The rank $r$, as a key hyper-parameter, plays an important role in the performance of LoRA, Adapter, LoTR, PISSA, and our proposed LoRA-PT method. To assess its impact, we randomly selected five samples from the EADC-ADNI dataset as the training set, with the remaining samples used for testing. Each method was evaluated under different rank settings, where $r \in \{1, 2, 4, 8, 16, 32\}$. Table 2 lists the Dice scores for methods that incorporate rank as a tunable parameter, including LoRA, Adapter, LoTR, PISSA, and LoRA-PT. As listed in the table, the optimal ranks were determined to be $r = 32$ for LoRA and Adapter, $r = 2$ for LoTR and PISSA, and $r = 1$ for LoRA-PT. These optimal values were used in all subsequent experiments unless otherwise noted.

Table 2. Dice (%) scores (mean±std) for segmentation results using different fine-tuning methods at various ranks on the EADC-ADNI dataset.

| Method | $r$=1 | $r$=2 | $r$=4 | $r$=8 | $r$=16 | $r$=32 |
|---|---|---|---|---|---|---|
| LoRA | 84.28±4.00 | 84.00±4.25 | 83.64±5.19 | 84.06±4.44 | 84.10±4.33 | **84.35±3.74** |
| Adapter | 83.58±5.76 | 83.80±5.16 | 83.91±5.21 | 83.49±5.47 | 83.74±4.56 | **84.08±5.28** |
| LoTR | 83.83±4.37 | **84.05±3.85** | 83.85±4.27 | 83.63±4.84 | 83.67±5.24 | 83.50±4.80 |
| PISSA | 84.16±3.43 | **84.25±3.53** | 83.91±3.80 | 83.68±3.72 | 83.91±3.78 | 83.86±3.65 |
| LoRA-PT | **84.47±3.40** | 83.72±3.77 | 83.83±3.77 | 83.61±3.90 | 83.84±3.75 | 83.69±3.83 |

*Note*: Higher Dice values indicate better segmentation performance. The best result for each method is highlighted in bold.

### *5.2. Comparison with state-of-the-art methods*

We conducted fine-tuning experiments under limited data conditions by randomly selecting five or ten samples from each of the three hippocampus datasets, using the remaining samples for testing. To enhance the robustness and reliability of the evaluation, we repeated the sampling process three times for each dataset, which effectively increased the size of the test set. Tables 3 and 4 list the segmentation performance of various methods under the five- and ten-sample training configurations, respectively. The proposed LoRA-PT method showed strong segmentation performance even with extremely limited training data. With only five training samples, the Dice score improved by 0.57%, 0.72%, and 2.34%, and the HD95 reduced by 0.15 mm, 0.02 mm, and 0.75 mm on the EADC-ADNI, LPBA40, and HFH datasets, respectively, compared to full tuning. When using ten training samples, LoRA-PT achieved Dice improvements of 0.44%, 0.72%, and 0.34%, and HD95 reductions of 0.42 mm, 0.19 mm, and 0.19 mm on the corresponding datasets, respectively.

Across all three datasets, LoRA-PT consistently outperformed the original LoRA method as well as the recently proposed LoTR and PISSA methods. These performance gains were statistically significant in most cases, as confirmed by the Wilcoxon signed-rank test. Furthermore, its robust reductions in HD95 across datasets validated the effectiveness and reliability of LoRA-PT under data-scarce conditions.

Table 3. Segmentation results on three hippocampus datasets, evaluated using Dice (%) and HD95 (mm) values. For each dataset, *five samples* were randomly selected as the training set, and the remaining samples were used for testing. This process was repeated three times, and the results listed in the table are the mean and standard deviation of the test data across all three repetitions, enhancing the reliability of the evaluation.

| Method | EADC-ADNI | | LPBA40 | | HFH | |
|---|---|---|---|---|---|---|
| | Dice↑ | HD95↓ | Dice↑ | HD95↓ | Dice↑ | HD95↓ |
| Full tuning | 84.07±3.41* | 5.24±1.97 | 81.61±3.39* | 6.68±1.58 | 79.10±6.67# | 5.81±1.73* |
| Linear-probing | 83.40±3.63* | **5.05±1.36** | 81.68±2.95* | 6.75±1.47 | 80.08±3.60* | 5.37±1.40* |
| LoRA | 83.81±3.51* | 5.22±1.74# | 81.54±3.01* | 6.82±1.60# | 80.51±4.85 | 5.35±1.48 |
| Adapter | 84.13±3.81* | 5.24±3.80 | 81.32±3.11* | 6.79±1.49 | 79.67±6.97# | 6.31±4.78# |
| SSF | 83.63±3.73* | 5.13±1.94 | 81.25±3.42* | 6.82±1.46 | 80.23±5.58* | 5.80±3.21* |
| LoTR | 84.14±3.67* | 5.09±1.66 | 81.68±3.14* | 6.81±1.51 | 80.25±4.84* | 5.56±2.58# |
| PISSA | 83.65±3.58* | 5.21±3.37 | 81.88±2.86* | 6.82±1.61 | 79.22±7.57* | 6.34±4.85* |
| LoRA-PT | **84.64±3.11** | 5.09±1.52 | **82.33±3.21** | **6.66±1.65** | **81.44±3.60** | **5.06±1.76** |

*Note:* Bold indicates the best result. * and # denote statistically significant improvements of LoRA-PT over the corresponding method based on the Wilcoxon signed-rank test (* for $p < 0.01$; # for $0.01 < p < 0.05$).

Table 4. Segmentation results on three hippocampus datasets, evaluated using Dice (%) and HD95 (mm) values. For each dataset, *ten samples* were randomly selected as the training set, and the remaining samples were used for testing. This process was repeated three times, and the results listed in the table are the mean and standard deviation of the test data across all three repetitions, enhancing the reliability of the evaluation.

| Method | EADC-ADNI | | LPBA40 | | HFH | |
|---|---|---|---|---|---|---|
| | Dice↑ | HD95↓ | Dice↑ | HD95↓ | Dice↑ | HD95↓ |
| Full tuning | 85.72±3.46* | 5.01±2.90* | 82.61±2.93* | 6.28±1.80 | 83.08±3.84 | 5.21±1.47 |
| Linear-probing | 85.22±3.20* | 4.74±1.21* | 82.52±2.92* | 6.19±1.69 | 82.86±4.02# | 5.19±1.59 |
| LoRA | 85.67±2.88* | 4.86±1.41* | 82.78±2.55* | 6.36±1.66 | 82.39±4.10* | 5.50±1.72* |
| Adapter | 85.49±2.90* | 4.83±1.28* | 82.66±2.62* | 6.29±1.66 | 82.92±3.81 | 5.23±1.91 |
| SSF | 85.12±3.22* | 4.90±1.37* | 81.57±2.97* | 6.58±1.59* | 81.86±4.02* | 5.24±1.70 |
| LoTR | 85.45±3.22* | 4.90±4.43# | 82.97±2.49# | 6.31±1.65 | 82.96±4.52 | 5.30±1.63 |
| PISSA | 85.98±2.81* | 4.68±1.26* | 82.94±2.55 | 6.22±1.69 | 83.03±4.17# | **4.97±1.47** |
| LoRA-PT | **86.16±2.83** | **4.59±1.32** | **83.33±2.71** | **6.09±1.59** | **83.42±3.98** | 5.02±1.65 |

*Note:* Bold indicates the best result. * and # denote statistically significant improvements of LoRA-PT over the corresponding method based on the Wilcoxon signed-rank test (* for $p < 0.01$; # for $0.01 < p < 0.05$).

We randomly selected one representative segmentation result from each of the three datasets for qualitative visualization, as shown in Fig. 3. Each row presents results generated by different fine-tuning methods. For each case, both 2D slice views and 3D surface renderings are shown. Blue regions indicate agreement between the automatic segmentation and the ground truth, while pink regions highlight discrepancies. As shown in the figure, the proposed LoRA-PT method produced segmentation results more closely aligned with the ground truth compared to the other methods.

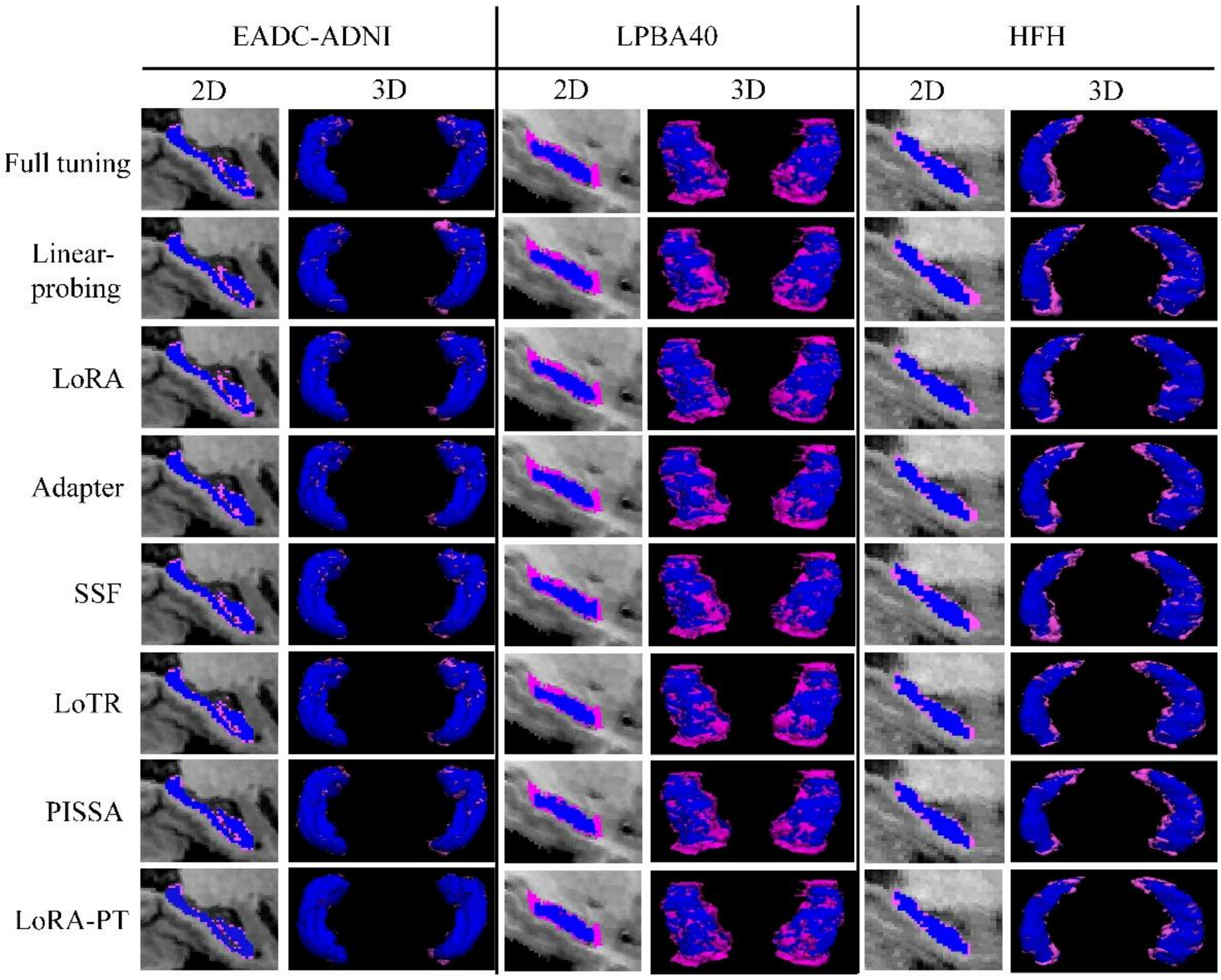

Fig. 3. Segmentation results using ten training samples. One representative result from each of the three datasets is shown. For each case, both 2D slice images and 3D surface renderings are provided. Blue regions indicate overlap between the automatic segmentation and the ground truth, while pink regions highlight areas of disagreement.

### 5.3. *Efficiency analysis*

We evaluated the computational efficiency of LoRA-PT by comparing several key metrics across various fine-tuning methods, including the number of trainable parameters, floating point operations (FLOPs) for additional modules in each transformer layer, GPU memory consumption, and training time per epoch. The comparison was conducted using five training samples. As listed in Table 5, LoRA-PT required only 2.84 million (M) trainable parameters, which is comparable to other PEFT methods such as SSF, LoTR, and PISSA, accounting for just 3.16% of the parameters required for full tuning. Additionally, LoRA-PT's FLOPs for the additional modules is 15.93 million (M), reflecting its relatively low computational cost.

Training memory usage and per-epoch training time were measured on an NVIDIA GeForce RTX 4090D GPU, with training time averaged over 1000 epochs to ensure stability. Among all methods evaluated, LoRA-PT achieved the lowest memory consumption at 11.15 gigabytes (GB) and the fastest training speed at 510 ms per epoch. Compared to full tuning, this represents a 39.04% reduction in memory usage and a 25.44% reduction in training time, highlighting its computational efficiency. These advantages are clearly shown in Fig. 4, which shows a scatter plot comparing memory consumption and training time per epoch across all methods. LoRA-PT, highlighted with a gold star, appeared in the bottom-left corner of the plot, indicating superior efficiency in both dimensions. PISSA also performed well, ranking just behind LoRA-PT. In contrast, full tuning appeared in the top-right corner, reflecting the highest computational cost. These results demonstrated that LoRA-PT substantially reduced training resource requirements while maintaining competitive segmentation performance, making it a promising solution for medical image analysis in resource-constrained environments.

Table 5. Comparison of trainable parameters, FLOPs of additional modules, GPU memory usage, and training time per epoch across methods. ΔM and ΔT denote the relative reductions in memory usage and training time compared to full tuning. Best results are shown in bold.

| Method | Trainable params (M) | Relative trainable params (%) | FLOPs of additional modules (M) | Memory (GB) | ΔM (%) | Training time per epoch (ms) | ΔT (%) |
|---|---|---|---|---|---|---|---|
| Full tuning | 90.01 | 100 | / | 18.29 | / | 684 | / |
| Linear-probing | 59.35 | 65.94 | / | 16.42 | -10.22 | 640 | -6.43 |
| LoRA | 7.40 | 8.22 | **9.67** | 15.90 | -13.07 | 562 | -17.84 |
| Adapter | 7.99 | 8.88 | 13.52 | 16.08 | -12.08 | 593 | -13.30 |
| SSF | 2.88 | 3.20 | 104.60 | 15.60 | -14.71 | 612 | -10.53 |
| LoTR | **2.70** | **3.00** | 35.43 | 17.24 | -5.74 | 602 | -11.99 |
| PISSA | 2.97 | 3.30 | 28.39 | 12.96 | -29.14 | 524 | -23.39 |
| LoRA-PT | 2.84 | 3.16 | 15.93 | **11.15** | **-39.04** | **510** | **-25.44** |

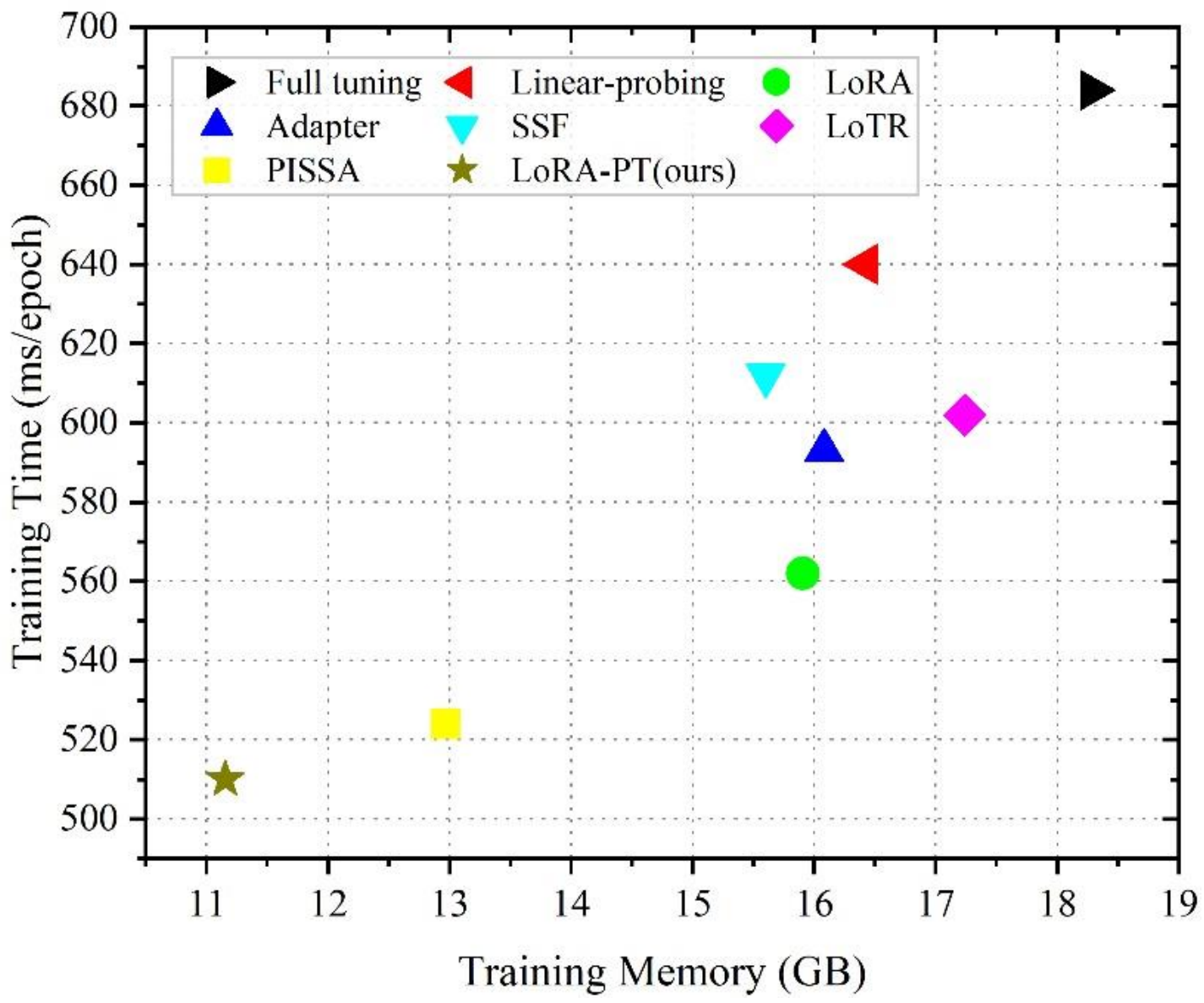

Fig. 4. Comparison of training time and GPU memory usage across different fine-tuning methods, evaluated on an NVIDIA GeForce RTX 4090D GPU. Training time is averaged over 1,000 epochs to ensure stability.

### *5.4. Ablation studies*

We conducted an ablation study to investigate the contributions of key components in LoRA-PT, focusing on three aspects: the effect of fine-tuning different transformer modules, the impact of selecting different singular components, and the influence of varying the t-SVD rank.

### *5.4.1. Fine-tuning of transformer components*

Table 6 lists the results of an ablation study on the impact of fine-tuning different transformer components in LoRA-PT, conducted on the EADC-ADNI dataset. Specifically, we evaluated the performance when selectively fine-tuning components including $W_q, W_k, W_v, W_o, W_{up}$, and $W_{down}$. The evaluation was based on five randomly selected training samples, with the remaining samples used for testing. Segmentation performance was measured using the Dice coefficient and HD95, with the best results highlighted in bold. The results indicate that fine-tuning all components achieves the highest Dice score and an acceptable HD95 value, ranking third among the four configurations. These results suggest that while optimal Dice performance is achieved, there may still be room for further improvement in boundary-level precision, as reflected by HD95.

Table 6. Ablation study of fine-tuning different transformer components in LoRA-PT on the EADC-ADNI dataset. *Five samples* were randomly selected as the training set, with the remaining samples used for testing. The Dice (%) and HD95 (mm) scores are reported as mean±std. The best results are highlighted in bold.

| $W_q, W_k, W_v$ | $W_o$ | $W_{up}$ | $W_{down}$ | #Parameters (M)↓ | Dice↑ | HD95↓ |
|---|---|---|---|---|---|---|
| √ | × | × | × | **2.73** | 81.88±4.89 | 5.43±5.43 |
| √ | √ | × | × | 2.75 | 82.42±5.17 | 5.21±1.38 |
| √ | √ | √ | × | 2.80 | 83.90±3.35 | **4.85±1.60** |
| √ | √ | √ | √ | 2.84 | **84.47±3.40** | 5.31±1.51 |

### *5.4.2. Impact of singular component selection*

To evaluate the effectiveness of the singular component selection strategy in LoRA-PT, we conducted an ablation study comparing three variants that initialized the trainable low-rank tensors with different subsets of tensor singular values and their corresponding singular vectors. Specifically, we tested the following strategies: initializing the trainable low-rank tensors with minor, medium, or principal singular components. The experiments were conducted on three public hippocampus segmentation datasets: EADC-ADNI, LPBA40, and HFH. As listed in Table 7, the principal variant consistently achieved the highest Dice scores and the lowest HD95 values across all datasets. In contrast, the medium and minor variants exhibited significantly lower performance.

These results confirm that the principal singular components encode the most informative and transferable directions in the parameter space. Focusing updates on these dominant components enables more effective adaptation, whereas modifying lower-magnitude components leads to diminished performance. This finding reinforces the core design of LoRA-PT, which concentrates parameter updates within the most relevant subspace derived from the pre-trained model. By avoiding unnecessary changes in less informative directions, this approach enhances generalization and training stability, especially under the limited-data scenarios commonly encountered in medical image segmentation.

Table 7. Comparison of singular component update strategies in LoRA-PT across three datasets. Dice (%) and HD95 (mm) values are reported (mean±std) for each strategy. For each dataset, *five samples* were randomly selected for training, and the remaining samples were used for testing. The best result for each metric and dataset is highlighted in bold.

| Method | EADC-ADNI | | LPBA40 | | HFH | |
|---|---|---|---|---|---|---|
| | Dice↑ | HD95↓ | Dice↑ | HD95↓ | Dice↑ | HD95↓ |
| Minor | 83.76±3.67 | 5.38±2.29 | 80.61±3.26 | 6.73±1.18 | 79.54±6.82 | 6.76±4.31 |
| Medium | 84.07±3.87 | 5.65±1.68 | 81.04±3.11 | 6.54±1.90 | 80.13±5.12 | 6.64±2.89 |
| Principal | **84.47±3.40** | **5.31±1.51** | **81.83±3.41** | **6.50±1.79** | **80.70±4.79** | **5.62±2.50** |

*5.4.3. Impact of t-SVD rank settings*

Table 8 lists the impact of t-SVD rank settings on segmentation performance across multiple datasets. The rank $r$ varied from 1 to 32, and the mean and standard deviation of the Dice coefficient were reported for each configuration. The training and testing protocols remained consistent, with five samples randomly selected for training and the remaining samples used for testing. The best result for each dataset is highlighted in bold. On the EADC-ADNI dataset, setting $r$=1 yielded the highest Dice score, indicating that a lower rank was more effective at capturing the key structural information within the data. Increasing the rank did not lead to further performance improvements. Similar trends were observed in the other datasets, where the optimal rank was also $r$=1. As listed in Table 8, segmentation performance remained relatively stable across different rank settings, except for the HFH dataset, where performance noticeably declined at $r$=8.

These results further validated the robustness of LoRA-PT and its ability to maintain high segmentation performance across different datasets and rank configurations. The consistent optimality at low ranks also underscores the method's efficiency in representing essential information with minimal parameter overhead.

Table 8. Dice (%) scores (mean±std) of segmentation results from the LoRA-PT method under different t-SVD rank settings across multiple datasets. For each dataset, *five samples* were randomly selected for training, and the remaining samples were used for testing. The best results are highlighted in bold.

| Dataset | $r=1$ | $r=2$ | $r=4$ | $r=8$ | $r=16$ | $r=32$ |
|---|---|---|---|---|---|---|
| EADC-ADNI | **84.47±3.40** | 83.72±3.77 | 83.83±3.77 | 83.61±3.90 | 83.84±3.75 | 83.69±3.83 |
| LPBA40 | **81.83±3.41** | 81.25±3.07 | 81.10±3.15 | 80.97±3.27 | 81.16±3.01 | 81.18±3.01 |
| HFH | **80.70±4.79** | 80.20±5.32 | 79.70±5.61 | 77.47±11.04 | 79.99±5.51 | 79.67±6.40 |

## 6. Discussion

### *6.1. Theoretical advantages of tensor-based methods in LoRA-PT*

The experimental results demonstrated that LoRA-PT consistently outperformed traditional matrix-based PEFT methods such as LoRA and PISSA across multiple datasets. This performance improvement can be theoretically attributed to the enhanced representational capacity of tensor-based modeling. While matrix-based methods use low-rank decompositions to reduce the number of trainable parameters, they are inherently limited to capturing pairwise correlations between rows and columns within individual parameter matrices. Such simplifications frequently fail to capture the rich, multi-dimensional interactions present in modern deep neural networks.

In contrast, LoRA-PT leverages t-SVD to construct a low-rank approximation in a higher-order parameter space. Specifically, it concatenates parameter matrices from different network layers along a new mode to form a third-order tensor and applies t-SVD to extract dominant components. This approach enables LoRA-PT to capture cross-layer dependencies that matrix-based methods are structurally unable to model. The t-product operation inherent in t-SVD further facilitates efficient computation by operating along the third mode (i.e., the layer dimension), allowing structured information exchange across layers while maintaining parameter efficiency. This modeling capability is especially important in medical image segmentation, where deep networks often rely on spatial, channel-wise, and layer-wise feature interactions to capture fine anatomical details. By adapting parameters within a multi-dimensional subspace that preserves these structural properties, LoRA-PT achieves more robust generalization under limited-data conditions, which are common in clinical settings.

In addition, the rank sensitivity analysis reveals that LoRA-PT achieved competitive performance even under extremely low-rank settings (e.g., rank 1), suggesting that a small number of dominant components were sufficient to capture the most transferable knowledge from the pre-trained model. This further highlights the efficiency and stability of the proposed method. These insights collectively underscore the theoretical foundations of LoRA-PT and provide a compelling explanation for its empirical success. By enabling expressive and efficient adaptation using tensor-based modeling, LoRA-PT represents a promising direction for advancing fine-tuning techniques in medical image analysis.

### *6.2. Explainability analysis of LoRA-PT*

To gain deeper insight into how LoRA-PT influences model behavior, we conducted an interpretability analysis using the LeGrad method [54], a technique specifically designed for vision transformers. LeGrad generates spatially detailed, class-relevant heatmaps by computing the gradients of attention maps with respect to the target prediction at each transformer layer. The encoder of the UNETR model consisted of 12 stacked transformer layers. To examine the evolution of attention across different levels of abstraction, we visualized three representative layers (Layers 3, 6, and 9), as shown in Fig. 5. The final column presents the original image with hippocampus annotations for reference.

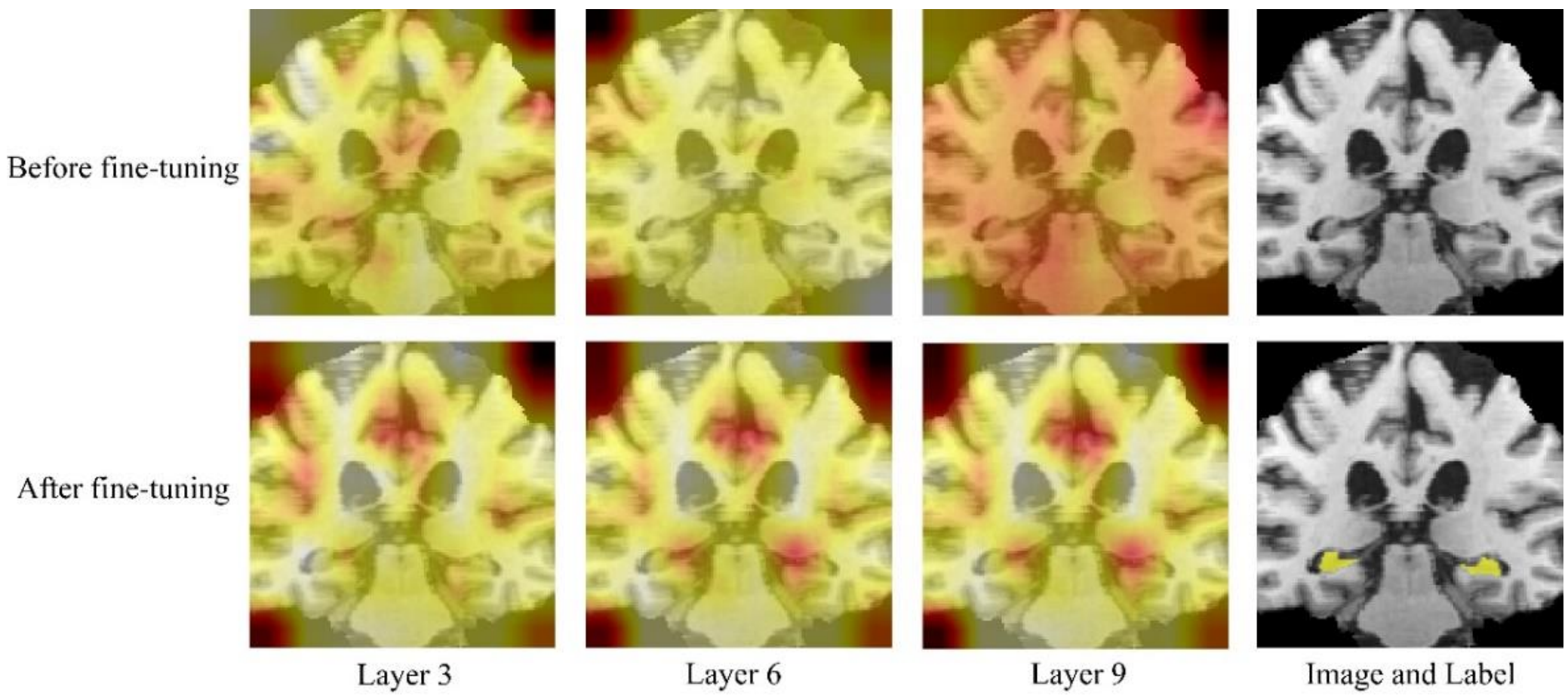

Fig. 5. Attention heatmaps generated by LeGrad from Layers 3, 6, and 9 of the UNETR encoder, showing changes in attention patterns before and after fine-tuning with LoRA-PT. The final column displays the original image with hippocampus annotations. After fine-tuning, attention in the deeper layers becomes more concentrated around the hippocampal region, indicating improved model performance in localizing task-relevant anatomical structures.

Before fine-tuning, attention maps from all layers were diffusely distributed, showing minimal focus on the hippocampal region. After fine-tuning with LoRA-PT, a clear shift in attention patterns was observed. In particular, deeper layers (Layers 6 and 9) exhibited significantly more concentrated activation around the hippocampus, indicating improved anatomical localization and semantic sensitivity. In contrast, the shallower layer (Layer 3) maintained a more diffuse focus, consistent with its role in capturing low-level features. This hierarchical refinement of attention suggests that LoRA-PT effectively reshaped the model's internal representations, guiding deeper layers to focus more precisely on semantically and anatomically relevant structures.

### *6.3. Impact of training data size on fine-tuning methods*

We investigated how varying the size of the training set influenced the performance of different fine-tuning methods on the EADC-ADNI dataset. As shown in the left panel of Fig. 6, when the number of training samples was fewer than 20, LoRA-PT and PISSA consistently outperformed other PEFT methods in terms of Dice score. Notably, LoRA-PT achieved the highest Dice scores with only five and ten training samples, highlighting its strong generalization capability under data-scarce conditions. These results suggest that the tensor decomposition mechanism used in LoRA-PT effectively captured structural information shared across layers, thereby enhancing model robustness when training signals were limited.

Once the training set exceeded 20 samples, full tuning began to outperform all PEFT methods in terms of Dice score. This is expected, as full tuning optimizes all model parameters and thus benefits more from abundant training data. However, this advantage comes at the cost of significantly higher computational demand and a greater risk of overfitting in small-sample scenarios. Importantly, LoRA-PT remained competitive even in large-data settings, achieving Dice scores close to those of full tuning while updating only a small fraction of parameters. Furthermore, as the training set size increased, the performance gap among different methods gradually narrowed.

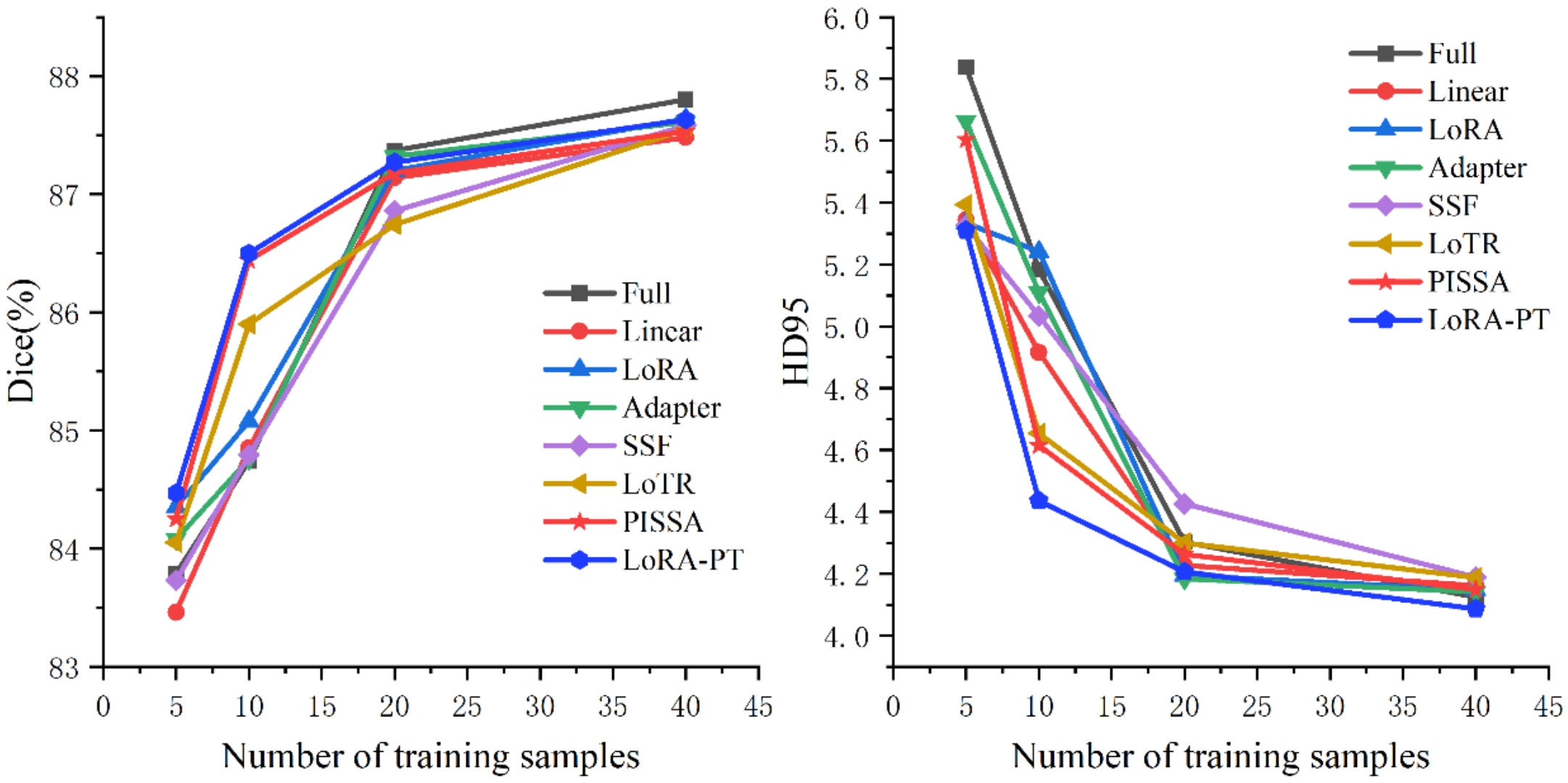

Fig. 6. Performance comparison of fine-tuning methods across different training sample sizes on the EADC-ADNI dataset, evaluated using Dice coefficient (left) and HD95 (right).

The right panel of Fig. 6 shows the HD95 trends. As the number of training samples increased, HD95 values decreased across all methods, indicating improved boundary segmentation accuracy. LoRA-PT consistently achieved the lowest HD95 across nearly all sample sizes, reflecting its superior ability to preserve anatomical boundaries. This not only reinforces its advantage in overall segmentation performance but also underscores its ability to retain clinically relevant structural details. These results demonstrate that LoRA-PT performed robustly in both low- and high-data regimes, further validating its practicality for real-world medical image segmentation tasks.

### *6.4. Impact of pre-training modality on LoRA-PT segmentation*

We examined the impact of pre-training modality on downstream segmentation performance through a series of comparative experiments. Specifically, the UNETR model was pre-trained using four MRI modalities from the BraTS2021 dataset (T1, T2, FLAIR, and T1ce). The resulting models were then fine-tuned using the LoRA-PT method on the EADC-ADNI dataset, which contained only T1-weighted images. Table 9 lists hippocampal segmentation results under two few-shot settings (five and ten training samples), evaluated using Dice score and HD95. Among all modalities, T1ce consistently achieved the highest Dice scores and the lowest HD95 values across both settings. This can likely be attributed to its superior image contrast and ability to preserve fine anatomical details, which allow the model to learn more discriminative and transferable features. Consequently, even when fine-tuned on a downstream task involving only T1-weighted images, the T1ce-pretrained model delivered superior performance.

Comparing results under the 5-sample and 10-sample training settings, we observed that as the number of fine-tuning samples increased, the performance gap between modalities gradually narrowed. This suggests that the influence of pre-training modality became less significant with larger training data, as the downstream task began to dominate the model's representational adaptation.

Table 9. Impact of pretraining with different MRI modalities on LoRA-PT hippocampal segmentation performance on the EADC-ADNI dataset. *Five* and *ten* samples were randomly selected as the training set, and results were evaluated using Dice (%) and HD95 (mm) values (mean±std).

| Modality | Five training samples | | Ten training samples | |
|---|---|---|---|---|
| | Dice↑ | HD95↓ | Dice↑ | HD95↓ |
| T1 | 77.97±5.25 | 6.96±4.79 | 85.37±3.35 | 5.92±5.91 |
| T2 | 75.71±7.06 | 7.00±5.50 | 84.07±5.54 | 5.45±4.53 |
| FLAIR | 78.54±5.50 | 6.91±5.35 | 84.61±4.78 | 5.51±4.76 |
| T1ce | **84.47±3.40** | **5.31±1.51** | **86.50±2.46** | **4.44±1.23** |

### *6.5. Model generalizability beyond hippocampus segmentation tasks*

To evaluate the generalizability of our proposed method beyond hippocampus segmentation, we conducted experiments on the UPENN-GBM dataset, which included MRI scans from 630 glioblastoma patients at the University of Pennsylvania Health System. This dataset provided a clinically relevant benchmark for brain tumor segmentation and is publicly available through the Cancer Imaging Archive [55]. Among the data, 147 cases have expert-annotated segmentation labels for key tumor sub-regions, such as the necrotic core, peritumoral edema, and enhancing tumor. In this study, we focused on the entire tumor region, using the contrast-enhanced T1 modality. To assess model performance in a low-supervision setting, we randomly selected 10% of the annotated samples for training and used the remaining 90% for testing.

Table 10 lists the segmentation results, evaluated using Dice score and HD95. The proposed LoRA-PT method achieved the highest Dice score (73.96%) and the lowest HD95 (27.20 mm), outperforming all other fine-tuning methods. Furthermore, its improvements over most baseline methods were statistically significant, as confirmed by the Wilcoxon signed-rank test. These results demonstrate that LoRA-PT generalized effectively beyond hippocampus segmentation, maintaining high accuracy and robustness on a more complex and heterogeneous tumor segmentation task.

Table 10. Performance comparison on the UPENN-GBM dataset, evaluated using Dice (%) and HD95 (mm) values (mean±std).

| Method | Dice↑ | HD95↓ | Method | Dice↑ | HD95↓ |
|---|---|---|---|---|---|
| Full tuning | 69.94±12.16* | 32.28±19.67# | SSF | 72.29±9.95* | 28.76±17.02# |
| Linear-probing | 72.42±10.79* | 29.40±18.81# | LoTR | 71.14±12.40* | 32.09±20.95* |
| LoRA | 72.51±11.17* | 29.80±17.90 | PISSA | 72.51±10.48* | 31.33±18.59* |
| Adapter | 73.46±9.92 | 33.36±20.74* | LoRA-PT | **73.96±9.75** | **27.20±17.15** |

*Note:* Bold indicates the best result. * and # denote statistically significant improvements of LoRA-PT over the corresponding method based on the Wilcoxon signed-rank test (* for $p < 0.01$, # for $0.01 < p < 0.05$).

### *6.6. Limitations*

Although LoRA-PT achieved promising segmentation performance, several limitations must be addressed to enhance its broader applicability. A primary concern is domain shift. While pre-training with modalities such as T1ce has shown strong transferability to T1-modality hippocampus segmentation, the model's ability to generalize across different MRI scanners and patient populations remains limited. Another challenge is the computational complexity introduced by the t-SVD. Although t-SVD enables efficient computation using the FFT, it still incurs additional overhead compared to the original LoRA method. As model size and task complexity increase, this overhead becomes more pronounced. Additionally, the current formulation of LoRA-PT is limited to third-order tensors, which is suitable for matrix-structured Transformer weights but not readily extendable to CNNs, where parameters naturally form higher-order tensors. This constraint may limit the flexibility of LoRA-PT in adapting to broader deep network architectures.

Finally, selecting the appropriate t-SVD rank is a critical factor influencing LoRA-PT's performance. In this study, we used grid search to determine the optimal rank. However, this method is neither efficient nor scalable when applying the model to new datasets or tasks. Manual tuning is time-consuming and may not consistently yield robust results, underscoring the need for more efficient and automated rank selection strategies.

### *6.7. Future studies*

In future studies, we will explore several key directions to enhance the scalability, adaptability, and theoretical grounding of LoRA-PT. First, we aim to develop multi-source learning frameworks and domain-invariant feature extraction techniques to better address domain shifts across institutions and imaging protocols. Second, we plan to investigate more efficient tensor decomposition strategies, such as randomized or low-memory approximations, and integrate them with model compression techniques, including quantization and pruning, to further reduce computational overhead. Third, we will explore dynamic rank adaptation mechanisms for t-SVD that are tailored to dataset-specific characteristics, thereby minimizing the requirement for manual tuning and improving model generalization.

Recent advances in tensor decomposition, such as mode-$k$ t-SVD, n-tubal rank modeling, basic-element tensor factorization, and multiplex transformed tensor decomposition [24, 27-29], offer new opportunities for modeling high-order tensor structures in a more flexible and robust manner. Building on these developments, we plan to extend LoRA-PT to CNNs for 3D medical image analysis. In this setting, the parameters of each convolutional layer are naturally represented as fifth-order tensors (e.g., output channels, input channels, kernel depth, height, and width), which can be stacked across layers to form a sixth-order tensor. This higher-order structure enables the integration of advanced tensor decomposition techniques into LoRA-PT, allowing for more expressive and structure-aware PEFT across diverse CNN architectures.

In addition, we will investigate the integration of zeroth-order optimization techniques with low-rank gradient tensor decomposition [56, 57]. These strategies have shown great potential in reducing memory usage and may further improve training efficiency in ultra-high-resolution medical imaging scenarios. Finally, to establish a stronger theoretical foundation, we plan to study the convergence properties of LoRA-PT under the t-SVD framework. This analysis will aid in bridging the gap between tensor optimization theory and the fine-tuning dynamics of deep neural networks, supporting the robustness and reliability of LoRA-PT in large-scale medical imaging applications.

## 7. Conclusions

In this study, we proposed LoRA-PT, a novel PEFT method based on t-SVD. Using LoRA-PT, we successfully transferred the UNETR model pre-trained on the BraTS2021 dataset to the hippocampus segmentation task and demonstrated its effectiveness across three public datasets. Experimental results show that LoRA-PT not only improved segmentation accuracy but also significantly reduced computational costs, making it a highly efficient and practical solution for medical image analysis, particularly in resource-constrained scenarios.

## CRediT authorship contribution statement

**Guanghua He**: Conceptualization, Data curation, Formal analysis, Investigation, Methodology, Writing - original draft. **Wangang Cheng**: Methodology, Software, Visualization. **Hancan Zhu**: Validation, Funding acquisition, Writing - review & editing. **Gaohang Yu**: Conceptualization, Formal analysis, Funding acquisition, Methodology, Project administration, Supervision.

## Declaration of competing interest

The authors declare no financial interests or personal relationships that could have influenced this study.

## Ethics statement

This study did not involve experiments on human or animal subjects. All analyses were conducted using publicly available, anonymized datasets in compliance with applicable laws and institutional guidelines. As no personally identifiable information was used and no new data were collected, ethical approval and informed consent were not required.

**Acknowledgments**

Gaohang Yu's work was supported in part by the National Natural Science Foundation of China (No. 12071104); Hancan Zhu's work was supported by the Humanities and Social Science Fund of the Ministry of Education of China (No. 23YJAZH232); and Guanghua He's work was supported by the Scientific Research Project of Shaoxing University (No. 20210038).